\documentclass[conference]{IEEEtran}
\IEEEoverridecommandlockouts
\usepackage{cite}
\usepackage{amsmath,amssymb,amsfonts, amsthm}
\newtheorem{theorem}{Theorem}
\usepackage{algorithm}
\usepackage{algpseudocode}
\usepackage{graphicx}
\usepackage{textcomp}
\usepackage{xcolor}
\usepackage{comment}
\usepackage{dblfloatfix} 
\usepackage{subcaption} 
\usepackage{balance}  

\def\BibTeX{{\rm B\kern-.05em{\sc i\kern-.025em b}\kern-.08em
    T\kern-.1667em\lower.7ex\hbox{E}\kern-.125emX}}
\begin{document}


\title{Constrained Multi-Objective Optimization for Automated Machine Learning\\
}


\author{\IEEEauthorblockN{
Steven Gardner,
Oleg Golovidov, 
Joshua Griffin,
Patrick Koch,
Wayne Thompson,
Brett Wujek and
Yan Xu
}
\IEEEauthorblockA{SAS Institute Inc.\\
North Carolina, USA\\
 \{Steven.Gardner,
Oleg.Golovidov,
Joshua.Griffin,
Patrick.Koch,
Wayne.Thompson,
Brett.Wujek,
Yan.Xu\}@sas.com
}}

\maketitle


\begin{abstract}
Automated machine learning has gained a lot of attention recently. 
Building and selecting the right machine learning models is often a multi-objective optimization problem.
General purpose machine learning software that simultaneously supports multiple objectives and constraints is scant,
though the potential benefits are great. 
In this work, we present a framework called Autotune that effectively 
handles multiple objectives and constraints that arise in machine learning problems. 
Autotune is built on a suite of derivative-free optimization methods, and 
utilizes multi-level parallelism in a distributed computing environment for automatically 
training, scoring, and selecting good models.
Incorporation of multiple objectives and constraints in the model exploration
and selection process provides the flexibility needed to satisfy trade-offs necessary
in practical machine learning applications.
Experimental results from standard multi-objective
optimization benchmark problems show that Autotune is very efficient in capturing Pareto fronts.
These benchmark results also show how adding constraints can guide the search to more 
promising regions of the solution space, ultimately producing more desirable Pareto fronts. 
Results from two real-world case studies demonstrate the effectiveness
of the constrained multi-objective optimization capability offered by Autotune.
\end{abstract}




\section{Introduction}

There has been increasing interest in automated machine learning (AutoML) for improving data scientists' productivity and 
reducing the cost of model building. A number of general or specialized AutoML systems 
have been developed ~\cite{googleAutoML,MSAutoML,TransmogrifAI,auto-sklearn,Auto-WEKA,h2o,jin2018efficient}, 
showing impressive results in creating good models with much less manual effort. 
Most of these systems only support a single objective, typically accuracy or error, to
assess and compare models during the automation process.   
However, building and selecting machine learning models is inherently a multi-objective optimization problem, 
in which trade-offs between accuracy, complexity, interpretability, fairness or inference speed are desired. 
There are a plethora of metrics for describing model performance~\cite{FeHeMo09,Po08} 
such as precision, recall, F1 score, AUC, informedness, markedness, and correlation to name a few.  
In general, each measure has an inherent bias~\cite{Po08} and
we typically expect data scientists to compare different performance measures when 
selecting the best models from a set of candidates. 
A data scientist might desire relatively accurate models 
but with minimal memory footprints and/or faster inference speed.
Alternatively, a data scientist might have business constraints that
are difficult to incorporate into the machine learning model 
training algorithm itself.  There could also be a number
of segments inherent within the data where it is important to have
comparable accuracy across all segments.  
When toggling between different performance measures and goals, 
what the data scientist is really doing is executing a manual multi-objective optimization. 
Arguably, they are mentally constructing a Pareto front and choosing the model 
that achieves the best compromise for their use case and criteria.  

 
It is considered fruitless to search for a single measure that perfectly captures
the multiple dimensions of interest in machine learning as shown in
Zitzler et. al~\cite{ZiLaThFoFo02} and paraphrased here:
 \begin{theorem}
 	In general, solution quality for the $m$-objective optimization problem
 	cannot be reduced to less than $m$ performance measures.
 \end{theorem}    
To emphasize this observation, we include a hypothetical example.  
Consider Matthews Correlation Coefficient (MCC) ~\cite{Matthews-MCC75}
that is considered a good metric to quantify performance
of the binary classification problem even when data is unbalanced:
\[
{\displaystyle \mathrm {MCC} ={\frac {\mathrm {TP} \times \mathrm {TN} -\mathrm {FP} \times \mathrm {FN} }{\sqrt {(\mathrm 
{TP} +\mathrm {FP} )(\mathrm {TP} +\mathrm {FN} )(\mathrm {TN} +\mathrm {FP} )(\mathrm {TN} +\mathrm {FN} )}}}}
\]
Now suppose we were to apply single objective optimization
and discover two models (model A and model B) with performances 
shown in Table \ref{ex_data_tab}.

\begin{table}[h]
\begin{center}
\begin{tabular}{|l|r|r|r|r|r|r|r|}\hline
	                 & TP    &  FP    & FN  & TN     &  ACC    &  MCC   &  FPR   \\\hline
\textbf{model A} & 900   & 500  & 100 & 8500  &  94.0\%  &   0.73   &  5.6\%  \\\hline
\textbf{model B} & 350   & 100  & 650 & 8900  &   92.5\% &   0.49   &  1.1\%  \\\hline
\end{tabular}
\end{center}
\caption[Two candidate models]%
{Performance of models A and B. TP is true positives; FP is false positives; FN is false negatives; TN is true negatives; ACC is accuracy; FPR is false positive rate. Compared to model B, model A has better MCC, but worse FPR.}
\label{ex_data_tab}
\end{table}

With MCC as the single objective to be maximized , an
optimization algorithm would discard model B in preference
for model A. 
However, the choice of which model is better depends entirely on context.
For instance, if this is a credit card fraud case, we might also be interested in  
reducing false positive rate (FPR) because false positives are very costly ~\cite{Bolt2018}. 
Thus, we would prefer to search around model B to attempt to improve MCC
while trying to maintain FPR. However, with unconstrained single objective optimization, 
this preference is difficult to enforce during the optimization process. 

One approach to addressing this problem is
aggregating multiple objectives into a
single objective, usually accomplished by some linear weighting
of the objectives. The main disadvantage of this approach is that many
separate optimizations with different weighting factors
need to be performed to examine the trade-offs among the objectives.
Another popular approach is multi-objective optimization ~\cite{JiSe08, ZiDeTh00}, 
which generates diverse multiple Pareto-optimal models to 
achieve a desired trade-off among various performance metrics and goals.
However, a potential drawback of pure multi-objective optimization is that
the corresponding algorithms are designed to determine the entire Pareto front when, in practice,
only part of the front may be desired.  For example, if considering
false negative rate and false positive rate together, the trivial models
that predict always negative and always positive could be part of the Pareto front.
It would be a waste of computational resources to train models to refine
such regions of the Pareto front.  Moreover, not all measures for assessing models 
can be easily formulated as objectives. 
Therefore, it can be very beneficial to guide model search to the desired area by using constraints. 



  
 In this work, we provide a constrained multi-objective optimization framework for 
 automated machine learning. This framework is built on a suite of
 derivative-free search methods and supports multiple objectives and linear or nonlinear constraints.  
 While the default search method works well in most settings,  
 the hybrid framework is extensible so that other desirable search methods
 can be incorporated easily in such a way that
 computing resources are shared to
 minimize and exploit inherent load imbalance. Moreover, redundant evaluations 
 are intercepted and handled seamlessly to avoid similar algorithms 
 within the hybrid strategy from performing redundant work. The approach works
 well on standard benchmark problems and shows promising results on
 real world applications. Our main contributions in this work are:
 
\begin{itemize}
\item To the best of our knowledge, this is the first general extensible constrained multi-objective optimization 
framework specifically designed for automated machine learning.  
\item The Autotune framework embraces the no-free-lunch theorem in that new and 
diverse search algorithms fit well in the existing framework
and may be added in a collaborative rather than a competitive manner,
permitting resource sharing and making completed evaluations available to 
all internal solvers that are capable of using them.  
\item By supporting general constraints,
we can aid users in focusing on specific segments of the Pareto front to save
computational time from models that are of little interest
to the user.  Further, in certain cases the multi-objective problem is
really a nonlinearly constrained problem in disguise; for example, 
one might wish only to optimize specificity and sensitivity while ensuring overall accuracy
does not degrade beyond a given threshold. The Autotune framework offers this flexibility.
\end{itemize}


\section{Related work}


Jin~\cite{JiSe08, Ji06} claims that machine learning is inherently a multi-objective
task and provides a compilation of various multi-objective applications
including feature extraction, accuracy, interpretability, and ensemble generation.
He et al.~\cite{HeHa18} use reinforcement learning to balance the trade-off between accuracy and compression of neural networks.  
The approach is sequential and not targeted toward the general multi-objective problem.
Asgari et al. ~\cite{TaKaMiHa17} apply a specialized evolutionary algorithm to optimize parameters
of an auto-encoder with respect to the two objectives: reconstruction error and classification error.
Loeckx~\cite{Lo15} stresses the need for multi-objective optimization in the context of machine learning
applied to structural and energetic properties of models, emphasizing that such an approach provides a gateway to hierarchy and abstraction.
A novel multi-objective evolutionary algorithm (ENORA)  was created to search for and select the
optimal feature subset in the context of a multi-class classification problem~\cite{Pa16}.
Shenfield and Rostami~\cite{ShRo17} apply an evolutionary algorithm that optimizes 
neural network weights, biases, and structures to simultaneously optimize both overall and individual class accuracy.  
In RapidMiner~\cite{RapidMiner2019}, an evolutionary framework is proposed where the user
may manually design the evolutionary algorithm using drag and drop features.  

A significant body of multi-objective research has been proposed in the context
of neural architecture search (NAS).  To simultaneously optimize accuracy
and inference speed, Kim et al.~\cite{KiReYuSe17} propose a multi-objective approach where neural architectures are 
encoded using integer variables and optimized using a customized evolutionary algorithm.
Elsken et al. ~\cite{ElMeHu18} develop a novel evolutionary algorithm (LEMONADE) to optimize
both accuracy and several model complexity measures including number of parameters. 
They propose a Lamarckian inheritance mechanism for warmstarting children networks with
parent network predictive performance. 
Dong et al. ~\cite{dong2018ppp} adopt progressive search to optimize for both device-related 
(inference speed and memory usage) and device-agnostic objectives (accuracy and model size).   
DVOLVER~\cite{GuMoLeFeFe19}, an evolutionary approach inspired by NSGA-II~\cite{DeAgPrMe00}, 
is created to find a family of convolutional neural networks with good accuracy and computational resource trade-offs. 

Multi-objective optimization in machine learning seems to favor evolutionary algorithms.
However, there have been enhancements made to many other derivative-free optimization approaches
that are appropriate and have complementary properties that, if combined, may create robust powerful hybrid
approaches. The derivative-free optimization community has been successfully handling these scenarios
in arguably similar if not identically complex and 
challenging conditions~\cite{DeAgPrMe00, AuSaZg08,CaShTa17,CuMa18}. For instance, 
inspired by direct-search methods, Cust{\'o}dio et al. ~\cite{CuMaVaVi11} propose a novel algorithm 
called direct multisearch for optimization problems with multiple black-box objectives. 
Deb and Sundar~\cite{DeSu06} combine a preference based strategy with an evolutionary multi-objective optimization 
methodology and demonstrate that a preferred set of solutions near a reference point can be found in parallel (instead of one solution).


\section{Constrained Multi-objective Optimization Framework}

Autotune is designed specifically to tune the hyperparameters and architectures of various machine learning model types including
decision trees, forests, gradient boosted trees, neural networks, support vector machines, 
factorization machines, Bayesian network classifiers, and more.
The tuning process utilizes customizable, hybrid strategies of search methods
and multi-level parallelism (for both training and tuning).  In this work, 
we focus on the two key features of Autotune: \textit{multiple objectives} and \textit{constraints}.

\begin{figure}[h]
   \includegraphics[width=.99\linewidth]{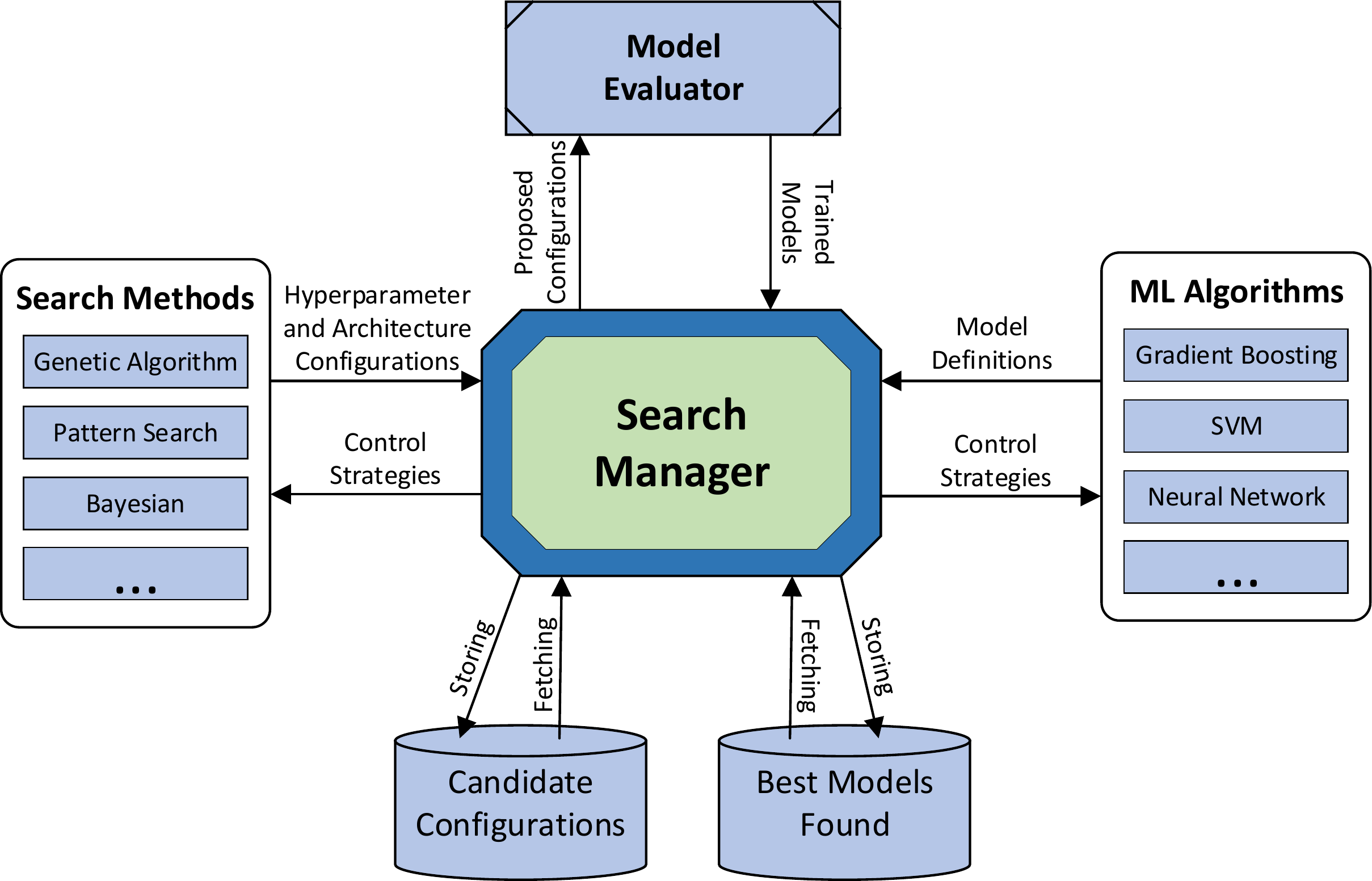}
    \caption[The Autotune Framework]
    {The Autotune framework. Machine learning algorithms provide detailed model definitions. 
   Search methods propose candidate configurations that are stored in a dedicated pool. 
    Model evaluator utilizes a distributed computing system to train and evaluate models.
    Search manager supervises the whole search and evaluation process, and  
    collects the best models found and other searching information.   }
    \label{fig2}
\end{figure}

The Autotune framework is shown in Figure \ref{fig2}. An extendable suite of
search methods (also called solvers) is driven by the search manager that controls concurrent
execution of the search methods. The search methods propose candidate configurations that 
are stored in a dedicated pool.  New search methods can easily be added to the framework. 
The model evaluator utilizes a distributed computing system to train and evaluate
candidate models. The search manager supervises the entire search and evaluation process and 
collects the best models found. The pseudocode in Algorithm \ref{alg:outeralg} provides a
high-level algorithmic view of the Autotune framework.


\begin{algorithm}
    \caption{Multi-objective constrained optimization in Autotune}
    \label{alg:outeralg}
    \begin{algorithmic}[1]
        \Require Population size $n_p$, and evaluation budget $n_b$.
        \Require Number of centers $n_c < n_p$ and initial step-size $\hat \Delta$.
        \Require Sufficient decrease criterion $\alpha \in (0,1)$. 
        \State Generate initial parent-points ${\mathcal P} $ using LHS with $|{\mathcal P}| = n_p$. 
        \State Evaluate ${\mathcal P}$ asynchronously in parallel.
        \State Populate reference cache-tree, ${\mathcal R},$ with unique points from ${\mathcal P}$.
        \State Associate each point $p \in {\mathcal P}$ with step $\Delta_p$ initialized to $\hat \Delta$.
        \State Let ${\mathcal F}$ denote current approximation of Pareto front.
        \While{$(|{\mathcal R}| \le n_b)$}
        \State Select ${\mathcal A} \subset {\mathcal F}$ for local search, such that $|{\mathcal A}| = n_c$.
        \For {$p \in {\mathcal A}$} \Comment{Search along compass directions}
        \State Set ${\mathcal T}_p = \{ \}$
        \For {$e_i \in I$}
        \State ${\mathcal T}_p = {\mathcal T}_p \cup \{p+\Delta_p e_i\}\cup \{p-\Delta_p e_i\}$
        \EndFor 
        \EndFor
        \State Generate child-points ${\mathcal C}$ via crossover and mutations on ${\mathcal P}$.
        \State Set ${\mathcal T} = {\mathcal C} \cup_{p \in {\mathcal A}}{\mathcal T}_p$. 
        \State Evaluate ${\mathcal T} \cap {\mathcal R}$ using fast tree-search look-up on ${\mathcal R}$.
        \State Project  ${\mathcal T} - {\mathcal R}$ to linear constraint manifold.
        \State Evaluate remaining ${\mathcal T} - {\mathcal R}$ asynchronously in parallel.
        \State Add unique points from  ${\mathcal T} - {\mathcal R}$ to cache-tree ${\mathcal R}$. 
        \State Update ${\mathcal P}$ with new generation ${\mathcal C}$ and initial step $\hat \Delta$.
        
        \For{$p \in {\mathcal A}$} 
        \If{$|{\mathcal T}_p\cap{\mathcal F}| > 0$}
        \State  Select new $p \in {\mathcal F}$ \Comment{Pattern search success}
        \Else
        \State Set $\Delta_p = \Delta_p/2$ \Comment{Pattern search failure}
        \EndIf
        \EndFor
        \EndWhile
    \end{algorithmic}
\end{algorithm} 

\subsection{Derivative-Free Optimization Strategy}

Autotune is able to perform optimization of
general nonlinear functions over both continuous and integer variables. The
functions do not need to be expressed in analytic closed form (i.e., black-box
integration is supported), and they can be non-smooth, discontinuous, and
computationally expensive to evaluate. Problem types can be single objective or
multi-objective.  The system is designed to run in either single machine mode or
distributed mode.

Because of the limited assumptions that are made about the objective 
and constraint functions, Autotune takes a parallel, hybrid, derivative-free
approach similar to those used in Taddy et al.~\cite{TaGr09}; Plantenga~\cite{Pl09}; Gray,
Fowler, and Griffin~\cite{GrFoGr10}; Griffin and Kolda~\cite{GrKo10}. Derivative-free methods
are effective whether or not derivatives are available, provided that the
number of variables is not too large (Gray and Fowler~\cite{GrFo11}). As a rule of thumb,
derivative-free algorithms are rarely applied to black-box optimization problems
that have more than 100 variables. The term ``black-box'' emphasizes that the
function is used only as a mapping operator and makes no implicit assumption
about the structure of the functions themselves. In contrast,
derivative-based algorithms commonly require the nonlinear objectives and
constraints to be continuous and smooth and to have an exploitable analytic
representation.

Autotune has the ability to simultaneously apply multiple
instances of global and local search algorithms in parallel. This ability streamlines
the process of needing to first apply a global algorithm in order to determine a
good starting point to initialize a local algorithm. For example, if the problem
is convex, a local algorithm should be sufficient, and the application of the
global algorithm would create unnecessary overhead. If the problem instead has
many local minima, failing to run a global search algorithm first could result
in an inferior solution. Rather than attempting to guess which paradigm is best,
the system simultaneously performs global and local searches while continuously
sharing computational resources and function evaluations. The resulting run time
and solution quality should be similar to having automatically selected the best
global and local search combination, given a suitable number of threads and
processors. Moreover, because information is shared among simultaneous searches, the robustness of this
hybrid approach can be increased over other hybrid combinations that simply use the
output of one algorithm to hot start the second algorithm. 

Autotune handles integer and categorical variables by using strategies and
concepts similar to those in Griffin et al.~\cite{GrFoGrHe11}. This approach can be viewed
as a genetic algorithm that includes an additional ``growth'' step, in which
selected points from the population are allotted a small fraction of the total
evaluation budget to improve their fitness score (that is, the objective
function value) by using local optimization over the continuous variables.

Execution of the system is iterative in its processing, with each iteration
repeating the following steps:
\begin{enumerate} 
	\item Acquire new points from the solvers
	\item   Evaluate each of those points by calling the appropriate black-box functions (model training and validation)
	\item   Return the evaluated point values (model assessment metrics) back to the solvers
\end{enumerate}

The search manager exchanges points with each solver in the list.
During this exchange, the solver receives back all the points that were
evaluated in the previous iteration.  Based upon those evaluated point values,
the solver generates a new set of points it wants evaluated and those new points
get passed to the search manager to be submitted for evaluation.  For
any solvers capable of ``cheating'', they may look at evaluated points that were
submitted by a different solver.  As a result, search methods can learn from
each other, discover new opportunities, and increase the overall robustness of
the system. 

To best utilize computing resources, Autotune supports multiple levels of parallelization ran simultaneously:
\begin{itemize} 
\item Each evaluation can use multiple threads and multiple worker nodes, and
\item Multiple evaluations can run concurrently
\end{itemize}
Evaluation sessions can be configured to minimize
the overlap of worker nodes but also allow resources to be shared.  
This design makes Autotune extremely powerful and capable of efficiently using 
compute grids of any size.



\subsection{Multi-Objective Optimization Approach}

When attempting to find the best machine learning model, 
it is very common to have several objectives. For instance, we might want to build models 
that maximize accuracy while also minimizing model size so that the models can
be deployed in mobile devices.
The desired result for such problems is usually not a single solution
but rather a range of solutions that we can use to identify an acceptable
compromise.  Ideally each solution represents a necessary compromise in the
sense that no single objective can be improved without causing at
least one remaining objective to deteriorate. The goal of Autotune in the multi-objective case is thus to
provide to the decision maker a set of solutions that represent the continuum of best-case scenarios.

Mathematically, we can define multi-objective optimization in terms of \emph{dominance}
and \emph{Pareto optimality}.
For a \textnormal{k}-objective minimizing optimization problem, a point (solution)
\textnormal{x} is \emph{dominated} by a point \textnormal{y} if
$f_i(x) \geq f_i(y)$ for all $i = 1,\ldots,k$ and $f_j(x) > f_j(y)$ for some $j = 1,\ldots,k$.

\begin{center}
\begin{figure}[h]
	\includegraphics[width=.80\linewidth]{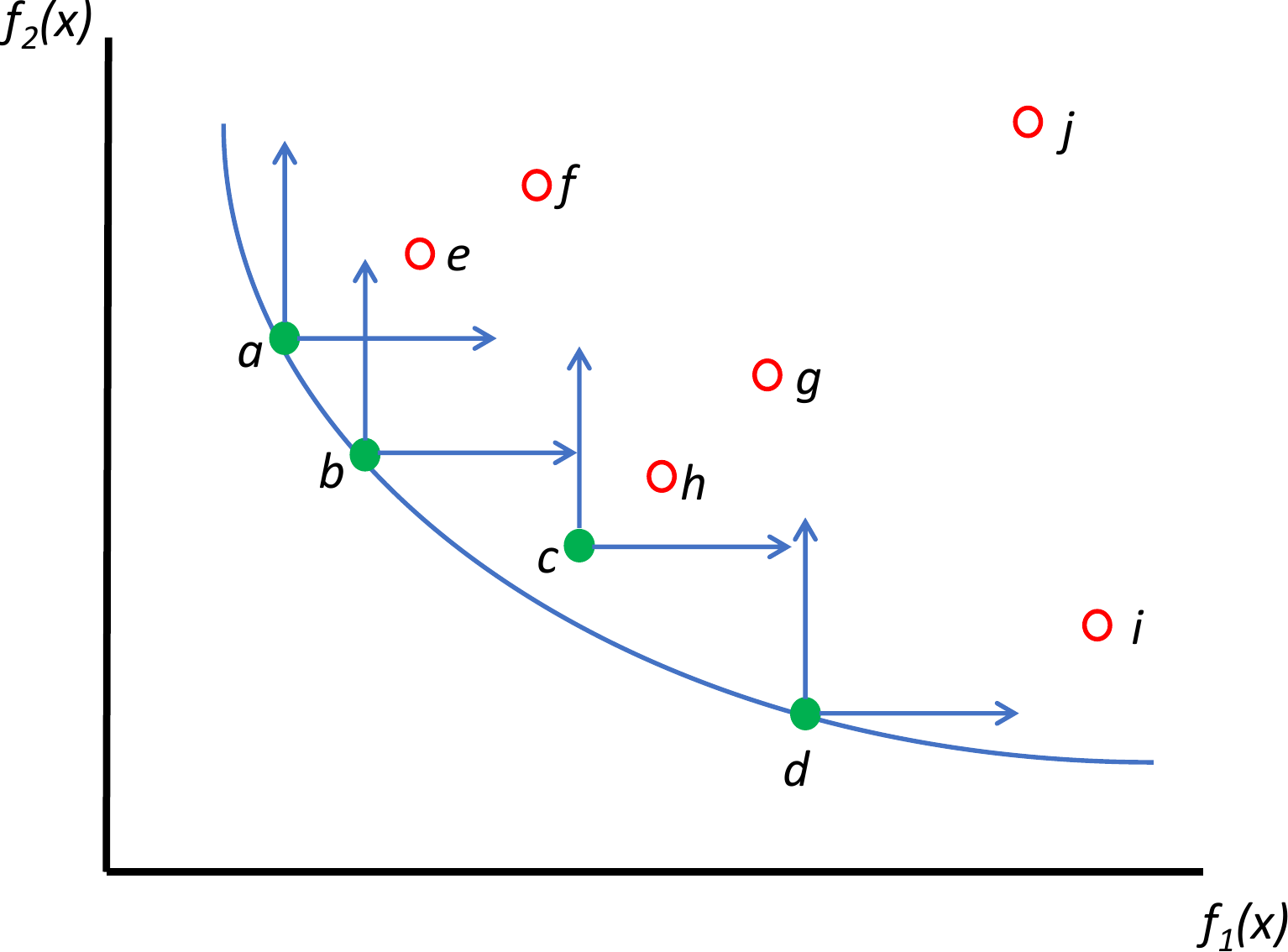}
		\caption{Example Pareto Front. $f_1(x)$ and $f_2(x)$ are two functions to be minimized. 
		Points \textnormal{$a$, $b$, $c$} and $d$ consist of the Pareto frontier found.}
	\label{pareto_front}
\end{figure}
\end{center}

A Pareto front contains only nondominated solutions.
In Figure \ref{pareto_front}, a Pareto front
is plotted with respect to minimization objectives $f_1(x)$
and $f_2(x)$ along with a corresponding population of 10
points that are plotted in the objective space. In this example,
point $a$ dominates $\{e,f,j\}$, $b$ dominates $\{e,f,g,j\}$,
$c$ dominates $\{g,h,j\}$, and $d$ dominates $\{i,j\}$.
Although no other point in the population dominates point $c$, 
it has not yet converged to the true Pareto front.
Thus there are points in a neighborhood of $c$ that have smaller values of
$f_1$ and $f_2$ that have not yet been identified.

\begin{figure*}[!b]
	\begin{subfigure}{.33\textwidth}
		\centering
		\includegraphics[width=.9\linewidth]{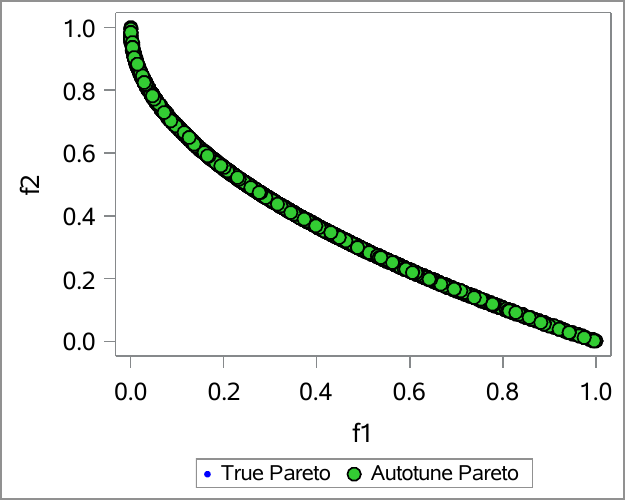}
		\caption{25,000 evals; no constraints}		
		\label{zdt1:25000}
	\end{subfigure}
	\begin{subfigure}{.33\textwidth}
		\centering
		\includegraphics[width=.9\linewidth]{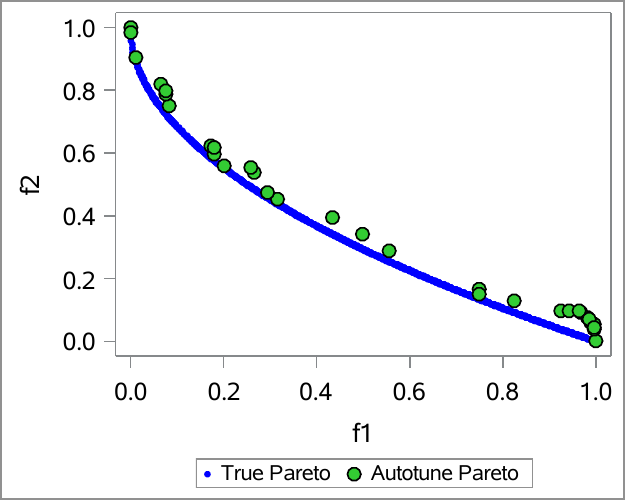}
		\caption{5000 evals; no constraints}		
		\label{zdt1:5000}
	\end{subfigure}%
	\begin{subfigure}{.33\textwidth}
		\centering
		\includegraphics[width=.9\linewidth]{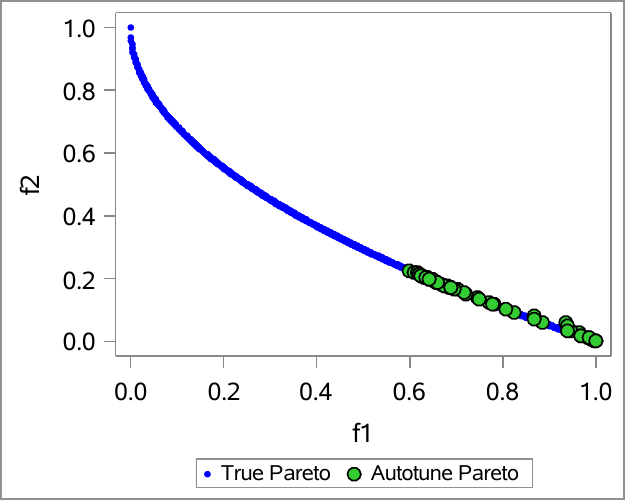}
		\caption{5000 evals; constraint $f_1 \geq 0.6$}	
		\label{zdt1:5000con}
	\end{subfigure}
	\caption{ZDT1 Benchmark Test Problem}
	\label{zdt1}
\end{figure*}

In the constrained case, a point $x$ is dominated by a point $y$ if
$\theta(x)  > \epsilon$ and $\theta(y) < \theta(x)$,
where $\theta(x)$ denotes the maximum constraint violation at point $x$ and
the feasibility tolerance is $\epsilon;$  thus feasibility takes precedence over objective function values.

Unlike common multi-objective optimization approaches 
that solely use metaheuristics~\cite{ElMeHu18,GuMoLeFeFe19,ShRo17}, 
the default approach employed by Autotune is a novel hybrid strategy that combines the global search emphasis of
metaheuristic ~\cite{Go89} with lesser known, but efficient, direct local search methods~\cite{GrKoLe08}.
The hybrid search strategy begins by creating 
a Latin Hypercube Sampling (LHS) of the search space.  This LHS is used as
the starting point for a Genetic
Algorithm (GA) to search the solution space for promising configurations.
GA's enable us to attack multi-objective
problems directly in order to evolve a set of Pareto-optimal solutions in one run of the
optimization process instead of solving multiple separate problems.
In addition, Autotune conducts local searches using a Generating
Set Search (GSS) algorithm in neighborhoods around nondominated points to
improve objective function values and reduce crowding distance. 
For measuring convergence,  Autotune uses a variation of the averaged Hausdorff distance ~\cite{schu_o:12} 
that is extended for general constraints.

\subsection{Constraint Handling}

In real-world use cases, it is common to encounter   
constraints that impose limits on the predictive models being used. 
For example, consider the context of the Internet of Things (IoT).
In the IoT setting, model size and inference speed are very important factors
as models are typically deployed to edge computing devices. 
If a model requires too much memory for storage or is very slow to score,
then it is not a good fit for edge computing. For mobile devices, models that
need many computations during inference will
consume too much power and should be avoided.  For these examples, it can be
extremely powerful to add constraints when picking a model.  The constraints
can be used 
to focus on the parts of the solution space that satisfy the business needs. 

Autotune uses different strategies to handle different types of constraints.
Linear constraints are handled by using both linear programming
and strategies similar to those in~\cite{grif_j:08},
where tangent directions to nearby constraints are constructed and used
as search directions.  In this case, trial points that violate the linear
constraints are first projected back to the feasible region before being
submitted for evaluation. Nonlinear constraints are handled by using
smooth merit functions \cite{grif_j:10b}.  Nonlinear constraint violations
are penalized with an L2-norm penalty term that is added to the objective
value. In the context of constrained
multi-objective optimization, when comparing points for domination,
a feasible point is always favored over an infeasible one.

\begin{figure*}
	\begin{subfigure}{.33\textwidth}
		\centering
		\includegraphics[width=.9\linewidth]{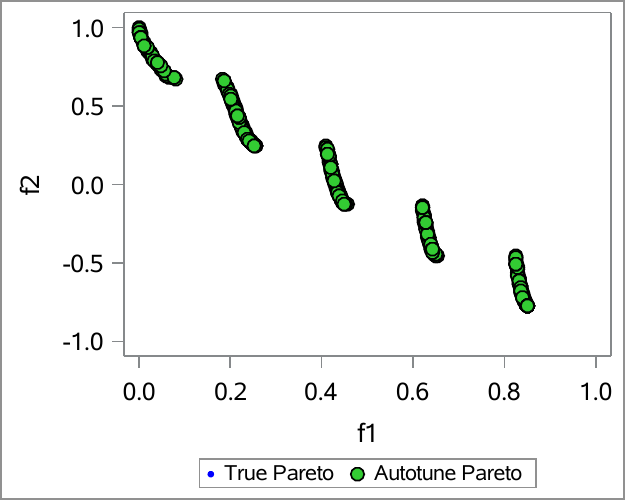}
		\caption{25,000 evals; no constraints}	
		\label{zdt3:25000}
	\end{subfigure}
	\begin{subfigure}{.33\textwidth}
		\centering
		\includegraphics[width=.9\linewidth]{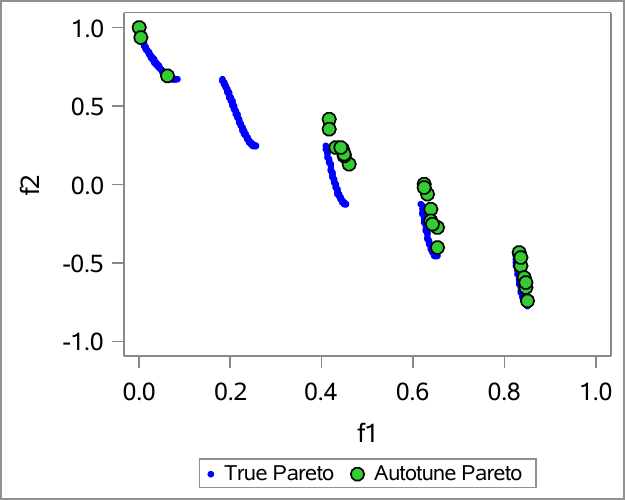}
		\caption{5000 evals; no constraints}
		\label{zdt3:5000}
	\end{subfigure}%
	\begin{subfigure}{.33\textwidth}
		\centering
		\includegraphics[width=.9\linewidth]{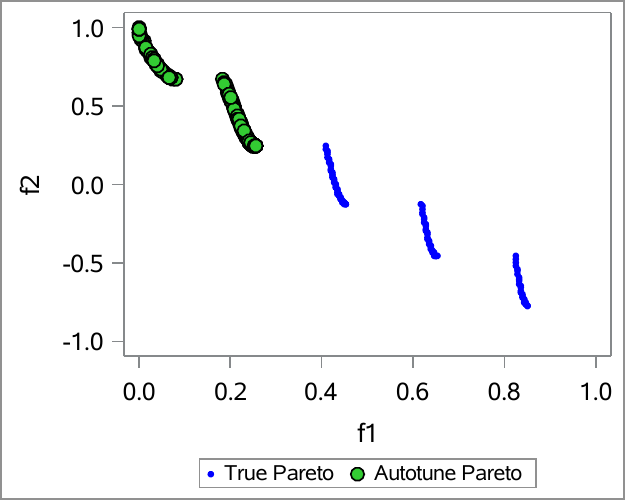}
		\caption{5000 evals; constraint $f_1 \leq 0.3$}	
		\label{zdt3:5000con}
	\end{subfigure}
	\caption{ZDT3 Benchmark Test Problem}
	\label{zdt1}
\end{figure*}


\section{Experimental Results}

While Autotune is designed specifically for automatically finding 
good machine learning models, the optimization
process that drives it is applicable to general optimization problems.
Therefore, to evaluate the performance of Autotune and its effectiveness at solving
multi-objective optimization problems, we conducted a benchmark experiment
by applying the Autotune system to a set of common multi-objective optimization
benchmark problems. We present a sampling of the results here for two of the 
benchmark problems: ZDT1 and ZDT3, taken
from \cite{ZiDeTh00}. For both of these problems, the true
Pareto front is known, which provides a basis for comparison.  

The mathematical formulation for ZDT1 is:
\[
f_1({x}) = x_1, \quad f_2({x})=g({x})(1-\sqrt{\frac{x_1}{g({x})}})
\]
\[
g({x})=1+\frac{9}{n-1}\sum_{i=2}^{n}x_i, \quad \forall x_i \in [0,1], n=30
\]

ZDT1 is a multi-objective optimization problem with two objectives ($f_1$, $f_2$) and 30 variables. 
Figure \ref{zdt1:25000} shows Autotune's results when running with a sufficiently
large evaluation budget of 25,000 evaluations.  The plot shows that 
Autotune has completely captured the true Pareto front and Autotune's Pareto markers
completely cover the true Pareto front. Many times in real-world use cases, evaluation budgets are 
limited due to time and cost.  Figure \ref{zdt1:5000} shows
Autotune's results when running with a limited evaluation budget of 5000 evaluations.
In this case, we can see that Autotune's approximation of the Pareto front isn't nearly as 
complete, and there are significant gaps when running with the limited evaluation budget.
Constraints can be added to the optimization
to focus the search to a particular region of the solution space. To demonstrate the power of constraints in the Autotune multi-objective
optimization framework, Figure \ref{zdt1:5000con} shows the results of re-running
Autotune against ZDT1, this time with a constraint specifying
that $f_1 \geq 0.6$. Again, Autotune was given a limited budget of 5000 evaluations. 
This plot clearly shows how adding the constraint has focused 
the optimization to that lower-right section of the solution space. This has 
allowed Autotune to capture a much better representation of the true Pareto front
in that region where $f_1 \geq 0.6$.

The mathematical formulation for ZDT3 is:
\[
f_1({x}) = x_1, \quad f_2({x})=g({x})(1-\sqrt{\frac{x_1}{g({x})}}-\frac{x_1}{g({x})}\sin{(10 \pi x_1)})
\]
\[
g({x})=1+\frac{9}{n-1}\sum_{i=2}^{n}x_i, \quad \forall x_i \in [0,1], n=30
\]

ZDT3 has two objectives ($f_1$, $f_2$) and 30 variables. Figure \ref{zdt3:25000} shows 
that Autotune is able to obtain the true Pareto front very well
when given a sufficiently large evaluation budget of 25,000 objective evaluations.
Figure \ref{zdt3:5000} shows Autotune's
results when running with a limited evaluation budget of 5000 objective evaluations.
Autotune struggles to find a complete representation of the Pareto front
when limited to 5000 evaluations.  In particular, the left side of the plot only shows
a few Pareto points that were found by Autotune.  Figure \ref{zdt3:5000con} shows 
the results with the same limited evaluation budget of 5000 objective evaluations but with 
an added constraint of $f_1 \leq 0.3$. The plot clearly shows Autotune was able to 
do a much better job of representing the Pareto front in that area of the solution space. 

This experiment demonstrates that Autotune correctly captures the Pareto fronts of the benchmark problems 
when given adequate evaluation budgets. By using constraints,  Autotune is able to significantly improve the search 
efficiency by focusing on the regions of the solution space that we are interested in. 

\section{Case Studies}

The case study data sets are much larger real world machine learning applications, using multi-objective optimization to tune a high quality predictive model.  The first data set comes from the Kaggle ‘Donors Choose’ challenge. 
The second data set is a sales leads data set. 
After a preliminary study of different model types, including logistic regression, decision trees, forests, and gradient boosted trees, the gradient boosted tree model type was selected for both case studies as the other model types all significantly underperformed.  Table \ref{case_tab1} presents the tuning hyperparameters of gradient boosted tree, their ranges, and default values.  

\begin{table}[h]
	\begin{center}
		\begin{tabular}{ | l | p{1cm} | p{1cm} | p{1cm} | }
			\hline
			\textbf{Hyperparameter} & \textbf{Lower} & \textbf{Default} & \textbf{Upper} \\ \hline
			Num Trees & \centering 100 & \centering 100 & \centering 500 \tabularnewline \hline
			Num Vars to Try & \centering 1 & \centering all & \centering all \tabularnewline \hline
			Learning Rate & \centering 0.01 & \centering 0.1 & \centering 1.0 \tabularnewline \hline
			Sampling Rate & \centering 0.1 & \centering 0.5 & \centering 1.0 \tabularnewline \hline
			Lasso & \centering 0.0 & \centering 0.0 & \centering 10.0 \tabularnewline \hline
			Ridge & \centering 0.0 & \centering 0.0 & \centering 10.0 \tabularnewline \hline
			Num Bins & \centering 20 & \centering 20 & \centering 50 \tabularnewline \hline
			Maximum Levels & \centering 2 & \centering 6 & \centering 7 \tabularnewline \hline
		\end{tabular}
	\end{center}
	\caption{Gradient Boosted Tree Hyperparameters}
	\label{case_tab1}
\end{table}

For both studies,  Autotune's default hybrid strategy that combines a LHS as 
the initial population with the GA and GSS algorithms is used. 
The population size used is 50 and the maximum number of iterations is 20. 
The tuning process is executed on a compute cluster containing 100 worker nodes.
Individual model training uses multiple worker nodes and multiple models are trained in parallel.  

\subsection{Donors Choose Data}
This case study involves building a model using data from the website DonorsChoose.org. This is a charity organization that provides a platform for teachers to request materials for projects. The business objective is to identify projects that are likely to attract donations based on the historical success of previous projects. Since DonorsChoose.org receives hundreds of thousands of proposals each year, automating the screening process and providing consistent vetting with a machine learning model allows volunteers to spend more time interacting with teachers to help develop successful projects. Properly classifying whether or not a project is “exciting” is a primary objective, but an important component of that is to minimize the number of projects improperly classified as exciting (false positives). This ensures
that valuable human resources are not wasted vetting projects that are likely to be unsuccessful.

The data includes 24 attributes describing the project, including:
\begin{itemize}
	\item the type of school (metro, charter, magnet, year-round, NLNS)
	\item school state/region
	\item average household income for the region
	\item grade level, subject, and focus area for the project
	\item teacher information
	\item various aspects of project cost
\end{itemize}

The data set contains 620,672 proposal records, of which roughly 18\% were ultimately considered worthy of a review by the volunteers. A binary variable labeling whether or not the project was ultimately considered “exciting” is used as the target for predictive modeling. The data set was partitioned into 70\% for training (434,470) and 30\% for validation (186,202) for tuning the gradient boosted tree predictive model. 


As mentioned in the study data set description, using misclassification rate as a single objective is not sufficient, and a successful predictive model is expected to also minimize the false positive rate. This makes the multi-objective optimization approach well suited for the study, with both misclassification rate and false positive rate (FPR) as the two objectives. It is unlikely that using any one of the more traditional machine learning metrics for tuning the models would produce the desired results.

The default gradient boosted tree model uses the default hyperparameter configuration listed in Table~\ref{case_tab1}.
Its confusion matrix is shown in Table \ref{donors_tab2}.  The default model predicts 5,562 false positives, a significant amount.
The FPR on the validation data set is 3.6\%. The overall misclassification rate on the validation set is high, around 15\%, and
needs to be improved, ideally while also improving FPR.

\begin{table}[h]
	\begin{center}
		\begin{tabular}{ | c |c | c | }
			\hline
			\textbf{Target} & \textbf{Predicted False} & \textbf{Predicted True} \\ \hline
			\textbf{False} & 146,956 & 5,562 \\ \hline
			\textbf{True} & 22,963 & 10,721 \\ \hline
		\end{tabular}
	\end{center}
	\caption{Confusion Matrix - Validation Data - Default Model}
	\label{donors_tab2}
\end{table}

\begin{figure}[h]
	\includegraphics[width=.99\linewidth]{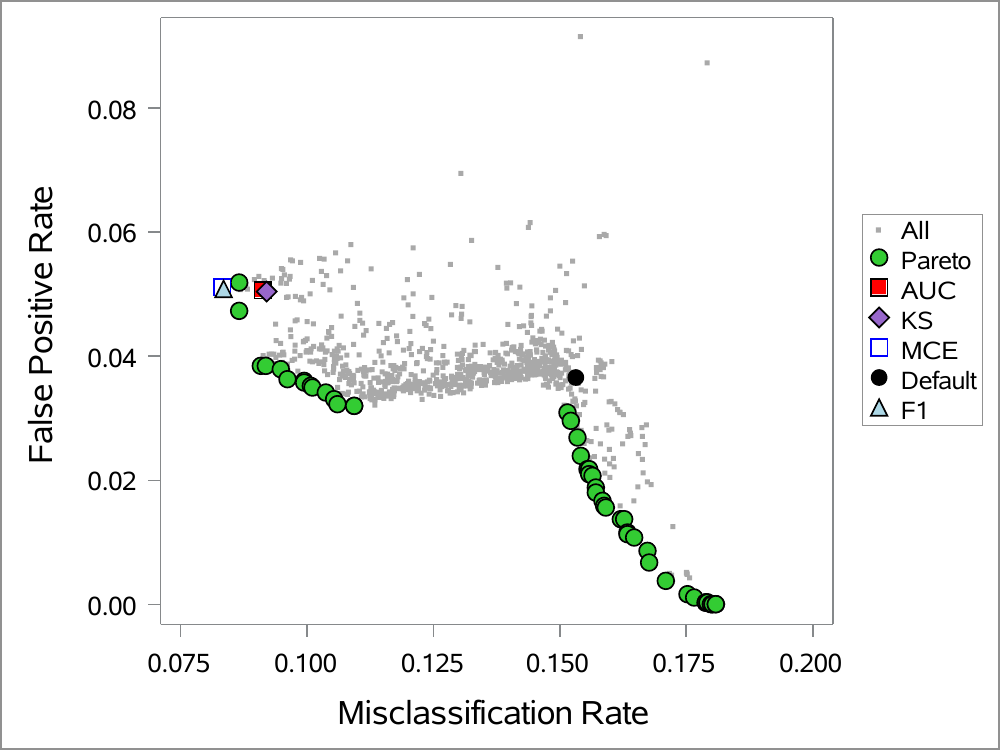}
	\caption{Donors Choose - All Evaluations}
	\label{donors_fig1}
\end{figure}

\begin{figure}[h]
	\includegraphics[width=.99\linewidth]{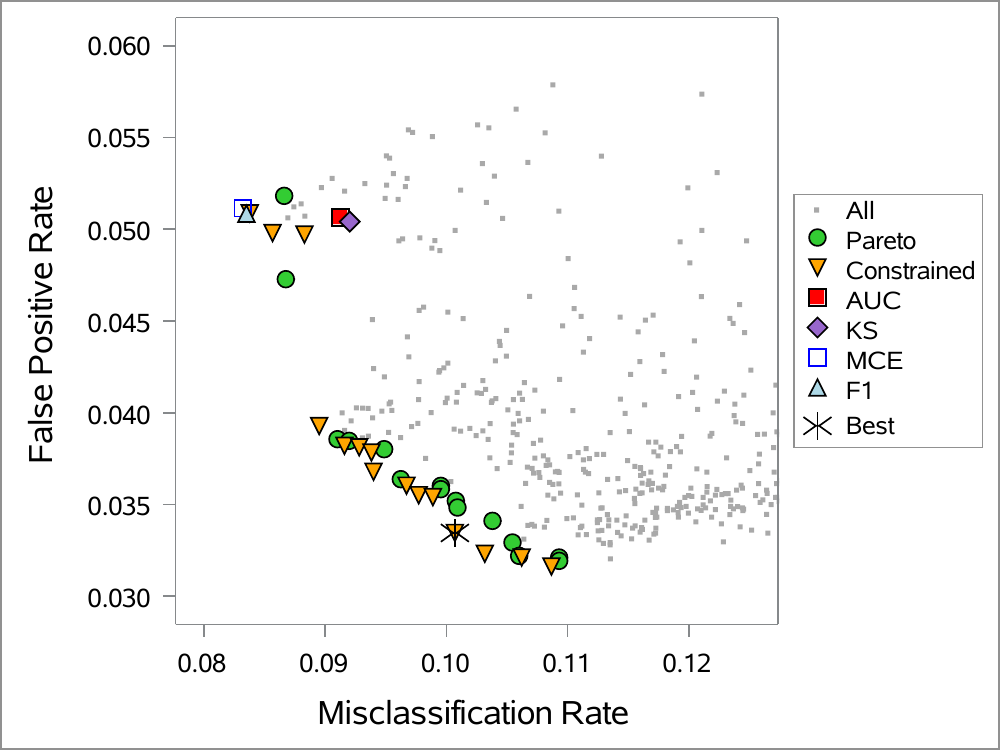}
	\caption{Donors Choose - Misclassification \textless 0.15}
	\label{donors_fig2}
\end{figure}

The multi-objective tuning results for the Donors Choose data set are shown in Figures \ref{donors_fig1} and \ref{donors_fig2}.  In Figure \ref{donors_fig1} the entire set of evaluated configurations is displayed, along with the default model and the generated Pareto front, trading off the minimization of misclassification on the x-axis and the minimization of the FPR on the y-axis. The entire cloud of points is split into two distinct branches, one branch trending towards a near zero FPR value, and another branch trending towards lower misclassification values, resulting in a split set of Pareto points. The default configuration appears to be a near equal compromise of the two objectives. 


Several other tuning runs were executed with various traditional metrics (AUC, KS, MCE and F1) as a single objective. The best model configurations for each of the runs are superimposed on Figures \ref{donors_fig1} and \ref{donors_fig2}. Nearly all of the single objective runs converged to similar values of misclassification and FPR. All of them sacrificed some FPR in the process, which is undesirable as defined by the conditions of this study.

While the near zero FPR values are appealing, the increase in the misclassification makes these configurations undesirable. It is more beneficial to look at models with both objectives reduced compared to the default model. Because of this, an additional tuning run was executed with an added constraint of misclassification \textless 0.15. The Pareto points for this tuning run are shown in Figure \ref{donors_fig2}. This figure shows a zoomed-in area around the points of interest and one of the Pareto points selected as the `Best' overall model. The confusion matrix for this 'Best' model is shown in Table \ref{donors_tab3}. The number of false positives reduced by 8\% (461) compared to the default
model but more importantly, the misclassification improved from 15\% to 10\%.


\begin{table}[h]
	\begin{center}
		\begin{tabular}{ | c | c | c | }
			\hline
			\textbf{Target} & \textbf{Predicted False} & \textbf{Predicted True} \\ \hline
			\textbf{False} & 147,417 & 5,101\\ \hline
			\textbf{True} & 13,650 & 20,034 \\ \hline
		\end{tabular}
	\end{center}
	\caption{Confusion Matrix - Validation Data - "Best" Model}
	\label{donors_tab3}
\end{table}

\subsection{Sales Leads Data}



Marketers often rely on machine learning models to accurately predict marketing actions and strategies that are most likely to succeed.
In this case study, we use a data set collected by the marketing department at SAS Institute Inc. 
A key goal of this study is to provide the sales team of the company with an updated list of quality 
candidate leads.
Supervised models are then built to identify and prioritize qualified leads across about 20 global regions. 
Machine learning qualifies leads by prioritizing known prospects and accounts based on their likelihood of acting.

The training data has about 200 candidate features through a four-year window. Web traffic data is a key feature category that includes page counts for several company websites as well as the referrer domain. Customer experience data such as the number of whitepapers downloaded, webcasts watched, and live events attended is also captured. A text analytics tool is used to standardize new features such as job function and department. Marketing based on business rules and actual outcomes labels the binary target for model training. The non-event (not a lead) is down sampled using stratified sampling to obtain a 10\% target event rate.  The data set contains 962,670 observations. For the tuning process, the observations were partitioned into 42\% for training (404,297), 28\% for validation (269,556), and 30\% for test (288,817). 

%

Purchase propensity models are very difficult to build due to the unbalanced nature of the training data. It is very important to deliver a scoring model that captures the event well yet minimizes false negatives so that sales opportunities are not overlooked. Typically with unbalanced data, overall misclassification rate is not the preferred measure of model quality. Here we investigate several model quality measures along with a multi-objective tuning strategy that incorporates both overall model accuracy and minimizing the false negative rate (FNR).  

The confusion matrix for a default gradient boosted tree model is shown in Table \ref{leads_tab4}. The default model predicts many more false negatives than false positives which is opposite from the desired scenario in this case – only 31\% of true positives are captured.



\begin{table}[h]
	\begin{center}
		\begin{tabular}{ | c | c | c | }
			\hline
			\textbf{Target} & \textbf{Predicted False} & \textbf{Predicted True} \\ \hline
			\textbf{False} & 276,718 & 1,193\\ \hline
			\textbf{True} & 7,542 & 3,364 \\ \hline
		\end{tabular}
	\end{center}
	\caption{Confusion Matrix - Holdout Data}
	\label{leads_tab4}
\end{table}

\begin{figure}[h]
	\includegraphics[width=.99\linewidth]{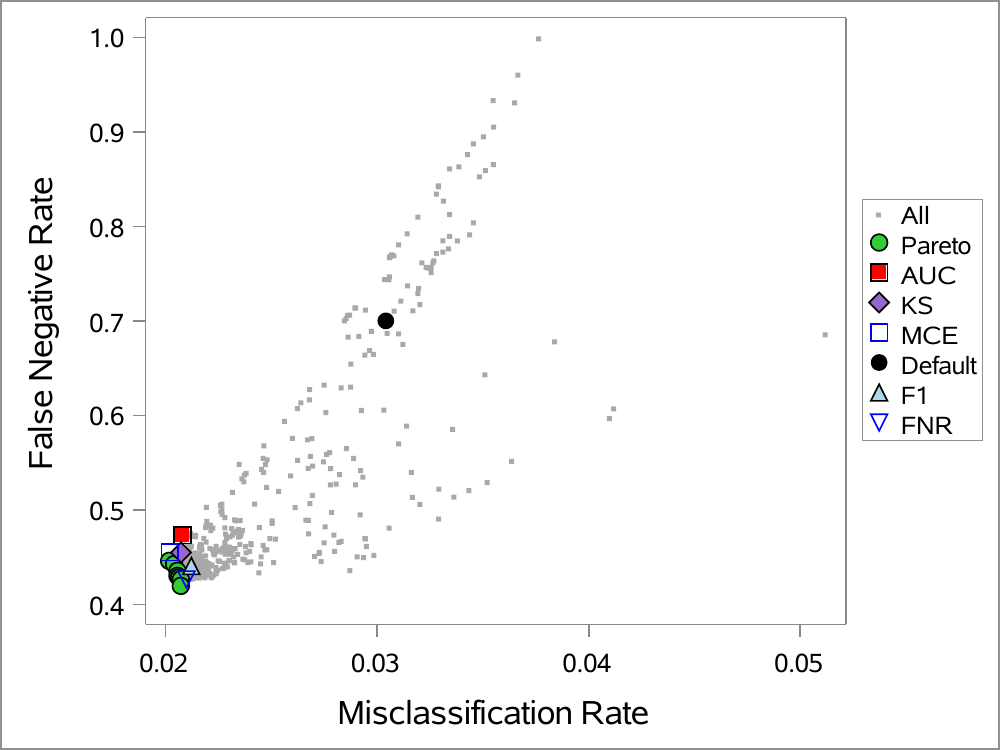}
	\caption{Leads Data Results - All Evaluations}
	\label{leads_fig1}
\end{figure}
\begin{figure}[h]
	\includegraphics[width=.99\linewidth]{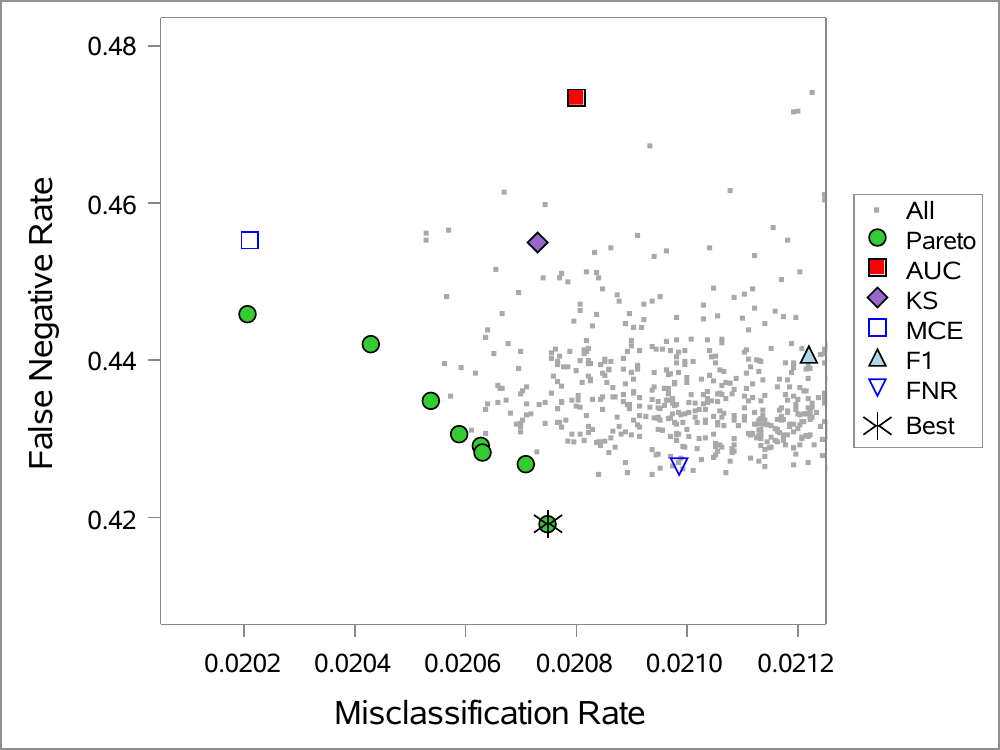}
	\caption{Leads Data - Pareto Front and Single Objective}
	\label{leads_fig2}
\end{figure}

The multi-objective tuning results for the leads data set are shown in Figures \ref{leads_fig1} and \ref{leads_fig2}. In Figure \ref{leads_fig1} the entire set of evaluated configurations is displayed, along with the default model and the generated Pareto front, trading off the minimization of misclassification on the x-axis and the minimization of the FNR on the y-axis.  The majority of the cloud of evaluations perform better than the default model, with respect to both overall misclassification and FNR. The Pareto front represents a set of trade-off solutions all of which are significantly better than the default model, cutting the FNR in half.

The Pareto front is shown in more detail in Figure \ref{leads_fig2}. It can be seen more clearly that the solution generated by maximizing only KS for this unbalanced data set, given the same evaluation budget, underperforms relative to the Pareto front of solutions. The overall misclassification of this solution is similar to that of the highest misclassification solution on the Pareto front and the FNR is higher than that of all solutions on the Pareto front. When the misclassification is minimized as a single objective tuning effort the misclassification is similar to the lowest misclassification solution on the Pareto front, but the FNR is higher. In review of the Pareto front, it is clear that the range of misclassification of the solutions is relatively small. If it is desirable to trade some false positives for a reduction of false negatives, an increase of over 300 sales leads can be obtained by sacrificing just 0.05\% in overall misclassification.

Constraints on both FNR and misclassification were applied in this problem in an attempt to identify more Pareto solutions with lower FNR.  However, since the Pareto front is very narrow in this case study, with both objectives gravitating towards the lower left in the solution space, no additional preferred Pareto solutions were identified by adding constraints. With very little trade-off between objectives observed after running multi-objective optimization, a final attempt to further reduce FNR is executed as a single objective constrained optimization problem. This result is shown in Figure \ref{leads_fig2}  which shows that when minimizing FNR directly as a single objective, we do not achieve results as desirable as those that were found when executing the multi-objective tuning process. The solution with the lowest FNR was chosen as the `Best' model and its confusion matrix is given in Table \ref{leads_tab5}.  The number of false negatives is reduced by 40\% (3007), compared to the default model. The FNR is 0.4343 on the holdout test data;  56.6\% of the true positive leads are captured, a significant improvement over 31\% with the default model.

\begin{table}[h]
	\begin{center}
		\begin{tabular}{ | c | c | c | }
			\hline
			\textbf{Target} & \textbf{Predicted False} & \textbf{Predicted True} \\ \hline
			\textbf{False} & 276,482 & 1,429\\ \hline
			\textbf{True} & 4,535 & 6,371 \\ \hline
		\end{tabular}
	\end{center}
	\caption{Confusion Matrix - Holdout Data - Lowest FNR}
	\label{leads_tab5}
\end{table}


\section{Conclusions}

Automation in machine learning improves model building efficiency and creates opportunities for more applications. 
This work extends the general framework Autotune by implementing two novel features: multi-objective optimization and constraints. 
With multi-objective optimization, instead of a single model, a set of models on a Pareto front
are produced. Then, the preferred model can be selected by balancing different objectives. 
Adding constraints is also important in the model tuning process. Constraints provide a way to enforce business restrictions or improve the search efficiency by pruning parts of the solution search space. 
The numerical experiments on benchmark problems demonstrate the effectiveness of our implementation
of multi-objective optimization and constraint handling. The two case studies we presented show
Autotune's ability to find models that appropriately balance multiple objectives while also adhering
to constraints. Future work to enhance Autotune includes simplifying the user's experience 
when choosing metrics for objectives and constraints.




\bibliographystyle{IEEEtran}
\balance 
\bibliography{IEEEabrv,multi-obj}

\end{document}